\documentclass[runningheads]{llncs}
\usepackage[T1]{fontenc}
\usepackage{graphicx}
\usepackage{booktabs}
\usepackage[misc]{ifsym}
\newcommand{\corr}{(\Letter)}
\usepackage[dvipsnames]{xcolor}
\usepackage{tcolorbox}
\usepackage[hidelinks]{hyperref}
\usepackage{amsmath}
\usepackage{amssymb}

\begin{document}

\title{Contagion on the Trading Floor: How Adversarial Signals Spread in Multi-Agent Trading Systems}

\toctitle{Contagion on the Trading Floor: How Adversarial Signals Spread in Multi-Agent Trading Systems}

\titlerunning{Contagion on the Trading Floor}

\author{Qi Rong Sua\inst{1} \and Junhao Dong\inst{1,3} \and Nguyen Duc Thai\inst{1} \and Yuqing Wen\inst{2} \and Cheston Tan\inst{3} \and Yew-Soon Ong\inst{1,3} \corr }
\tocauthor{Qi Rong Sua, Junhao Dong, Nguyen Duc Thai, Yuqing Wen, Cheston Tan, Yew-Soon Ong}
\authorrunning{Sua et al.}

\institute{ Nanyang Technological University, Singapore\\ \email{\{suaq0001,junhao003,ducthai001,asysong\}@ntu.edu.sg} \and National University of Singapore, Singapore\\ \email{e1351400@u.nus.edu} \and Centre for Frontier AI Research, IHPC, A*STAR, Singapore \\ \email{cheston\_tan@a-star.edu.sg}}

\maketitle              

\begin{abstract}
Multi-agent trading systems built on large language models (LLMs) are beginning to appear in quantitative finance, yet their robustness to adversarial inputs is largely unknown. We study the vulnerability of \emph{LLM} trading stacks to black-box, input-only attacks that enter solely via admissible social-media feeds. We introduce the \emph{Generic Multi-Agent Trading System} (GMATS), a framework that captures modern multi-agent trading architectures and instantiate a class of black-box poisoning attackers that treat an LLM as a post generator and inject budget-constrained, plausibly benign social-media content into the analyst’s evidence stream. We define contagion metrics that trace how adversarial content propagates through the stack, including belief-shift scores at analyst and coordinator layers and attack--clean deltas on standard backtest metrics. Experiments on a safe offline benchmark with historical market and social data show that even simple input-only attackers can materially degrade risk--return profiles, sharply reducing Sharpe ratios. At the same time, we find that suitably designed multi-agent topologies and coordinator prompts can dampen adversarial shocks and improve average robustness under identical poisoning budgets. Our code and dataset are available in this \href{https://github.com/Soqoro/GMATS}{\textcolor{RoyalBlue}{repository}}.
\end{abstract}

\section{Introduction}
The rapid rise of large language models (LLMs) has led to a corresponding surge in LLM-based agents and multi-agent systems \cite{autogen,camel,metagpt,chatdev,guo2024multiagents_survey}. In quantitative finance, a growing line of work has built multi-agent trading systems on top of LLM agents, reporting substantial gains in backtested performance metrics. These architectures typically decompose a trading desk into specialized agents (fundamental, sentiment, risk, execution) that communicate via natural language to produce trades \cite{tradingagents,tradinggroup,contesttrade,quantagent}. At the same time, recent results such as \textsc{AgentSmith} \cite{agentsmith} showing that a single adversarial input can jailbreak many multimodal LLM agents exponentially fast highlight how composition can \emph{amplify} rather than mitigate vulnerabilities. When agents reason about one another's outputs, small perturbations at one node may cascade into large changes in the final decision. 

This tension raises a natural question for emerging \emph{LLM-native} trading stacks:
\begin{tcolorbox}[
  colback=cyan!4,
  colframe=cyan!45!black,
  boxrule=0.45pt,
  arc=2pt,
  left=5pt,
  right=5pt,
  top=4pt,
  bottom=4pt
]
\emph{How vulnerable are multi-agent, LLM-only trading systems to input-only,
black-box attacks that enter solely through admissible evidence streams?}
\end{tcolorbox}

We present, to the best of our knowledge, the first systematic study of the robustness of multi-agent trading systems to such black-box, input-only adversaries. We focus on attacks that operate entirely through a realistic and important channel: equity-related social media. Among typical evidence streams (fundamentals, news, social), social media poses the largest adversarial surface  \cite{nam2025pumpdump,liu2024publicshortcampaigns,kogan2023socialmedianewsmanip}. It is high-volume, weakly curated, and easily seeded with misinformation \cite{pohl2024socialbotsfinance}, yet removing social-media analysts is often not realistic: sentiment and flow signals from social platforms are a major driver of performance in many equity strategies \cite{bollen2011twittermood,tetlock2007investorsentiment,antweiler2004talk,sprenger2014tweets}. To analyze these threats in a way that is comparable across architectures, we introduce the \emph{Generic Multi-Agent Trading System} (GMATS), a framework tailored for LLM-based multi-agent trading systems. GMATS captures a common pattern across existing systems: \emph{analyst} agents ingest environment data (fundamentals, news, social media) and produce textual insights; one or more \emph{coordinator} agents fuse these insights into a trading stance and critically evaluate them; optional \emph{controller} and \emph{executor} components enforce risk and translate stances into trades.

\paragraph{Contributions.}
\begin{enumerate}
    \item \textbf{GMATS: a unifying framework for LLM trading stacks.}
    We introduce the \emph{Generic Multi-Agent Trading System} (GMATS), a framework for LLM multi-agent trading systems and we prove that a broad class of existing trading agents can be compiled into this form.

    \item \textbf{Contagion metrics and evaluation protocol.}
    We define contagion metrics that track how adversarial content propagates through the stack, including belief-shift scores (BSS) at analyst and coordinator layers and attack--clean deltas on standard backtest metrics (returns, volatility, Sharpe). This yields a reusable protocol for stress-testing multi-agent LLM trading systems.

    \item \textbf{Dataset and reference implementation.}
    We use an offline one-year corpus of equity-related social-media posts (Twitter and Reddit) to drive the social analyst’s evidence stream. Dataset details and release information are provided in Section~\ref{sec:experiments}.

    \item \textbf{Empirical study of attacks and defenses.}
    We instantiate input-only poisoning attackers and quantify their impact on belief contagion and risk--return profiles under varying poisoning rates. We then compare alternative GMATS architectures including single-agent, multi-analyst, and two defense-oriented coordinator designs (wide bull--bear and a pairwise-triangulation coordinator) and show that coordinator-level structure can substantially dampen adversarial shocks and improve robustness under identical poisoning budgets.
\end{enumerate}


\paragraph{Related works.} Recent LLM-driven trading frameworks coordinate specialized agents (e.g., fundamental, sentiment, technical, risk) via debate and role decomposition to emulate trading desks. \emph{TradingAgents} formalizes this pattern with analyst teams feeding a decision agent under risk controls, while \emph{TradingGroup} adds structured self-reflection and an end-to-end data-synthesis loop to improve coordination and backtested performance. \emph{ContestTrade} introduces an internal contest mechanism with separate Data and Research teams whose outputs are continuously scored and filtered by real-market feedback, and \emph{QuantAgent} targets high-frequency trading with price-driven Indicator, Pattern, Trend, and Risk agents operating over short-horizon market signals. Together, these systems highlight the importance of inter-agent communication, ranking, and ingestion pipelines that GMATS abstracts and that our input-only attacker seeks to perturb \cite{tradingagents,tradinggroup,contesttrade,quantagent}.
However, these systems are typically evaluated only on backtested performance under benign data, leaving their robustness to adversarial evidence especially social media and the resulting \emph{contagion dynamics} across agents largely unexplored.
\section{Generic Multi-Agent Trading System (GMATS)}

\begin{figure}[t]
    \centering
    \includegraphics[width=\textwidth,trim=20 15 20 15,clip]{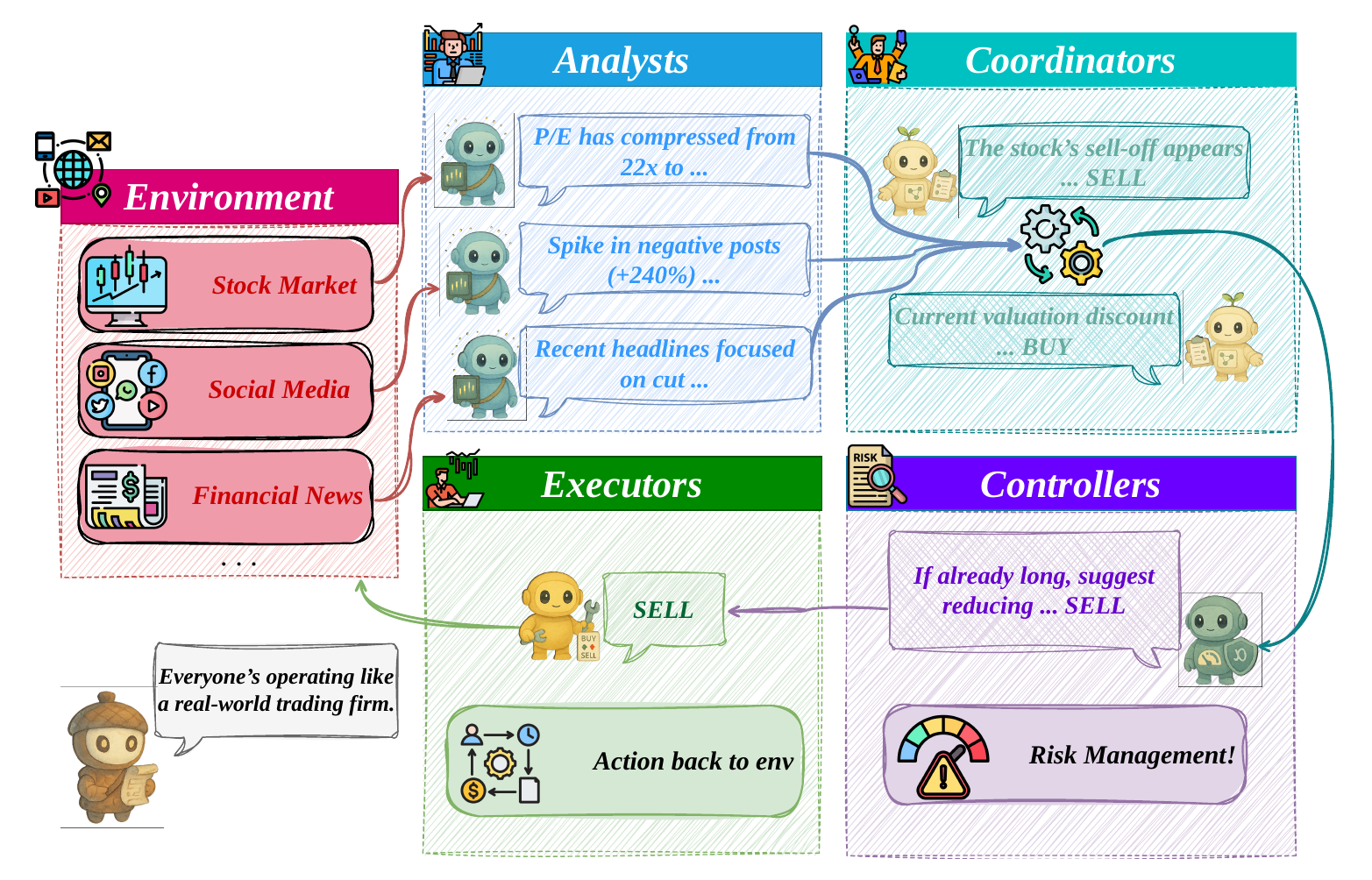}
    \caption{Architecture of the Generic Multi-Agent Trading System (GMATS).}
    \label{fig:gmats}
\end{figure}

\begin{definition}[GMATS]\label{def:gmats}
(Architecture in Figure~\ref{fig:gmats}.) Let $\mathcal{A}$ be the asset set ($|\mathcal{A}| = n$) and let $\mathbb{T}$ be a discrete time index set.
A \emph{GMATS} instance is the tuple
\[
G = \bigl(\mathcal{A}, A, C, K, X,\ \mathcal{S},\ \Sigma,\ \mathcal{D},\ \mathcal{L},\ \Lambda,\ \mathcal{E}\bigr),
\]
with the following components:
\begin{itemize}
\item \textbf{Agents.} Analysts $A$ ($|A|\!\ge\!1$), coordinators $C$ ($|C|\!\ge\!1$), controllers $K$ ($|K|\!\ge\!0$), executors $X$ ($|X|\!\ge\!0$).  
Agents may be LLM-based or programmatic.

\item \textbf{Typed messages and schema.}  
$\mathcal{S}$ is a finite set of message types.
A message is
\[
m = (\mathrm{id}, \mathrm{src}, \mathrm{dst}, \tau, \mathrm{payload}, \mathrm{ts}),
\]
with $\tau\in\mathcal{S}$ and timestamp $\mathrm{ts}\in\mathbb{T}$.  
Messages live on a shared bus; at round $t\in\mathbb{T}$, only messages with $\mathrm{ts}\le t$ may be consumed (point-in-time, PIT).

\item \textbf{Schedule.}  
$\Sigma$ specifies, for each round $t$, the order in which agents are activated together with any random seeds used in their calls (e.g., LLM sampling seeds).
Each round $t$ executes the pipeline $A \!\to\! C \!\to\! K \!\to\! X$, with micro-rounds permitted inside each stage.

\item \textbf{Data and environment.}  
$\mathcal{D} = \{\mathcal{D}_t\}_{t\in\mathbb{T}}$ denotes point-in-time data sources (market, news, social, fundamentals, etc).  
$\mathcal{E}$ is the execution environment with state space $\mathcal{X}$, which maps orders and current state to fills and next state:
\[
(\mathrm{fills}, \chi_{t+1}) = \mathcal{E}(o_t, \chi_t),\qquad \chi_t\in\mathcal{X}.
\]

\item \textbf{Coordination and control.}  
$\mathcal{L} = \{\mathcal{L}_t\}_{t\in\mathbb{T}}$ denotes coordinator logic that maps analyst insights and context to stances and draft orders, and $\Lambda = \{\Lambda_t\}_{t\in\mathbb{T}}$ denotes controller logic that enforces risk and constraint policies on these stances or orders.
We write $\xi_t$ for any additional context (e.g., portfolio summaries, constraints) supplied to coordinators and controllers at round $t$.
\end{itemize}
\end{definition}

\paragraph{Contracts (per round $t$).}
\begin{enumerate}
\item \textbf{Analysts ($A$).}  
Each $a\in A$ emits zero or more \emph{insight} messages $m^{(a)}_j$ with type $\tau\in\mathcal{S}$ and $\mathrm{ts}\le t$, computed from PIT data $\mathcal{D}_t$ and the inbox of $a$.

\item \textbf{Coordinators ($C$).}  
Consuming all analyst insights at round $t$, the coordinator logic $\mathcal{L}_t$ produces a \emph{stance}
\[
(s_t, \rho_t) = \mathcal{L}_t(\text{insights}_t, \xi_t), \qquad s_t\in\mathbb{R}^{n},
\]
with metadata $\rho_t$.  
Optionally, a draft order $o_t\in\mathcal{O}$ may be produced in place of $s_t$.

\item \textbf{Controllers ($K$).}  
The risk/constraints gate
\[
(\mathrm{ok}_t, o'_t) = \Lambda_t(z_t, \xi_t), \qquad z_t\in\{s_t, o_t\},\quad o'_t\in\mathcal{O},
\]
enforces limits (exposure, leverage, liquidity, CVaR/SL/TP, compliance) and yields an executable order~$o'_t$ (or $\mathrm{ok}_t=\mathrm{false}$ if no trade is allowed).

\item \textbf{Executors ($X$).}  
If $|X|>0$, an executor submits $o'_t$ to the environment:
\[
(\mathrm{fills}_t, \chi_{t+1}) = \mathcal{E}(o'_t, \chi_t).
\]
If $|X|=0$, $\mathcal{E}$ is called directly by the coordinator/controller as the final step.
\end{enumerate}

\paragraph{Orders space.}
Discrete orders live in
\[
\mathcal{O}_{\mathrm{disc}} = \{-1, 0, 1\}^{n},
\]
and continuous portfolio weights live in
\[
\mathcal{O}_{\mathrm{cont}} \subseteq \mathbb{R}^{n},
\]
e.g., with $\ell_1/\ell_\infty$ and leverage caps.
We write $\mathcal{O} = \mathcal{O}_{\mathrm{disc}} \cup \mathcal{O}_{\mathrm{cont}}$ for the overall order space.

\paragraph{Invariants.}
(i) \textbf{PIT:} all consumed evidence/messages satisfy $\mathrm{ts}\le t$;  
(ii) \textbf{Replayability:} each run emits a complete log of messages, prompts, LLM responses, and orders, so that the resulting trajectory can be replayed offline for the same $\Sigma$ without re-querying the LLMs;  
(iii) \textbf{Schema safety:} only well-typed messages ($\tau\in\mathcal{S}$) are routable.

\begin{theorem}[Expressivity of GMATS]\label{thm:gmats-expressivity}
Let $\mathcal{M}$ be a multi-agent trading pipeline that satisfies:
\begin{enumerate}
    \item \textbf{Typed message passing.} Agents in $\mathcal{M}$ communicate only via messages from a finite schema $\mathcal{S}_\mathcal{M}$, and each agent's behavior is a function of its inbox and point-in-time environment data.
    \item \textbf{Point-in-time discipline.} At round $t$, every agent in $\mathcal{M}$ may only consume messages and environment data with timestamps $\le t$.
    \item \textbf{Stage separation.} For each round $t$, the computation of trading decisions in $\mathcal{M}$ can be decomposed into four phases: (i) evidence ingestion, (ii) fusion and scoring, (iii) constraints and risk checks, and (iv) order execution.
\end{enumerate}
Then there exists a GMATS instance
\[
G = (\mathcal{A}, A, C, K, X, \mathcal{S}, \Sigma, \mathcal{D}, \mathcal{L}, \Lambda, \mathcal{E})
\]
such that, for any initial environment state and fixed random seeds, the sequence of executed orders produced by $\mathcal{M}$ is identical to the sequence produced by $G$. We provide the proof in supplementary materials.
\end{theorem}

\begin{corollary}[Compilation of existing frameworks]
TradingAgents, TradingGroup, ContestTrade, and QuantAgent all satisfy the conditions of Theorem~\ref{thm:gmats-expressivity} and can be compiled into GMATS by mapping their ingestion, fusion, risk, and execution components into the $A \to C \to K \to X$ pipeline and unrolling their internal coordination into micro-rounds of~$\Sigma$.
\end{corollary}

\section{Metrics}

\paragraph{Setup.}
We evaluate an \emph{attack} run against a \emph{clean} (no-poison) baseline over $T$ decision times and $N$ tradable assets.
Let $\mathcal{P}$ be the set of poisoned messages, and let $\operatorname{Top}_k(t)$ denote the $k$ highest-ranked items actually consumed by the social analyst at time $t$.
We define the poison-consumption rate
\begin{equation}
p_{\text{poison}} \;=\; \frac{1}{T} \sum_{t=1}^T \frac{\bigl|\operatorname{Top}_k(t) \cap \mathcal{P}\bigr|}{k},
\end{equation}
and in Experiments~A--C we vary the attacker’s budget until $p_{\text{poison}}$ hits target levels (e.g., $15\%, 30\%, \dots, 75\%$).
Unless stated otherwise, performance metrics are reported as \emph{attack minus clean}.

\subsection{Contagion Metrics}
\label{sec:metrics}
\paragraph{Belief Signal Shift (BSS).}
BSS measures how the attack systematically shifts the agent’s internal numeric beliefs, even before trades are placed.
Let $f_{i,t}^{\text{attack}}$ and $f_{i,t}^{\text{clean}}$ denote the forecast for asset $i$ at time $t$ under the attacked and clean runs, respectively.
We define the aggregate belief shift as
\begin{equation}
\mathrm{BSS}
= \frac{1}{TN}\sum_{t=1}^{T}\sum_{i=1}^{N}
   \bigl(f^{\text{attack}}_{i,t}-f^{\text{clean}}_{i,t}\bigr).
\end{equation}
A positive BSS indicates that, on average, forecasts become more bullish under attack, while a negative BSS indicates a net shift toward more bearish beliefs.
\subsection{Performance Metrics (FinSABER; attack--clean)}

\noindent\textbf{Protocol.}
We adopt the FinSABER performance protocol as described in \cite{li2025llmbasedfinancialinvestingstrategies}.
Let $\{r_t\}_{t=1}^{T}$ be the net (after-cost) portfolio returns at the chosen frequency (daily by default).
Define $A$ as the annualization factor ($A{=}252$ for daily).

\paragraph{Cumulative and annualized return.}
These metrics quantify how much the portfolio grows in total (CR) and what that growth corresponds to as an annualized rate (AR), under attack versus clean.
\begin{equation}
\mathrm{CR}=\prod_{t=1}^{T}(1+r_t)-1,\qquad
\mathrm{AR}=(1+\mathrm{CR})^{A/T}-1.
\end{equation}
\paragraph{Annualized volatility.}
Annualized volatility captures the scale of return fluctuations, i.e., how ``noisy'' or risky the strategy is over a year.
\begin{equation}
\mathrm{AV}=\sqrt{A}\,\operatorname{sd}(r_t).
\end{equation}

\paragraph{Sharpe ratio.}
Sharpe summarizes risk-adjusted performance using total volatility \cite{sharpe1966mutualfund}.
\begin{equation}
\mathrm{SR}=\frac{\sqrt{A}\,\operatorname{mean}(r_t)}{\operatorname{sd}(r_t)}.
\end{equation}

\paragraph{Delta reporting.}
To isolate the effect of the attack, we report performance as the difference between attack and clean runs.
For each metric $m\in\{\mathrm{CR},\mathrm{AV},\mathrm{SR}\}$, we compute
\begin{equation}
\Delta m = m^{\text{attack}} - m^{\text{clean}}.
\end{equation}

\section{Attacker Model}
\label{sec:attacker}
\paragraph{Attacker taxonomy}
We consider two simple but representative attack policies. Both respect the same daily budget $B$ on the single target asset. \textbf{random attacker} serves as a minimal baseline. On each day $t$ it spends its budget $B$ by generating posts with random sentiment directions, independent across posts and days (e.g., sampling bullish vs.\ bearish with equal probability). This attacker injects noise without targeting and any contagion it induces reflects a general sensitivity of GMATS to low-level social perturbations. The \textbf{always-negative attacker} is a simple one-sided stressor. On each day $t$ it spends its entire budget $B$ generating strongly bearish posts about the target asset. This policy approximates a persistent negative-narrative campaign (e.g., a ``short-and-distort'' style effort) that continuously pushes sentiment in the same direction. Together, these two attackers span a simple design space: untargeted, sign-symmetric noise (random) and untargeted but persistently one-sided pressure (always-negative). Our experiments in Section~\ref{sec:experiments} compare their induced contagion profiles under common budgets $B$ and varying \(p_{\text{poison}}\).

\paragraph{Scope and admission assumption.}
We measure \emph{downstream contagion conditioned on adversarial content being present in the social evidence stream}, rather than modeling how such content is created, promoted, or filtered. Because real trading stacks differ widely in post scoring/admission (heuristics, sentiment models, LLM filters), an end-to-end seeding model would confound results with system-specific details. We therefore assume attacker posts are \emph{admissible} and vary attack strength via the \emph{effective poison rate} \(p_{\text{poison}}\), defined as the fraction of \emph{consumed} (top-\(k\)) posts attributable to the attacker, which isolates susceptibility under a fixed retriever. This is realistic on open social channels where volume can surface some posts without retriever compromise \cite{ferrara2016socialbots,vykopal2024disinfo}, and where pipelines often upweight high-salience or strongly valenced content \cite{milli2025engagement,su2025corpuspoison,zou2025poisonedrag}. Thus \(p_{\text{poison}}\) serves as an abstract knob capturing posting intensity and any upstream amplification while keeping evaluation focused on what happens \emph{after} poisoned posts enter the system.

\section{Experiments}
\label{sec:experiments}

\subsection{Data and Setup}

We instantiate GMATS with three analyst roles (social, fundamentals, news), one coordinator, one risk controller, and one executor. All experiments use \texttt{gpt-4o-mini} and ran over the same historical window (\textbf{01 Nov 2021--01 Dec 2021}) for AAPL. We focus on a single ticker/month for controlled mechanistic analysis. Extending to more tickers or months primarily increases sample size (tightening uncertainty) but does not change the propagation mechanism captured by our contagion metrics under the same PIT evidence-admission and poisoning budget. We also release a one-year corpus of equity-related social media posts collated from tweets and reddit for the top 7 large-cap tech companies (AAPL, AMZN, GOOG, META, MSFT, TSLA, NFLX) spanning 30 Sept 2021 -- 29 Sept 2022. Each record provides a timestamp, tweet text, ticker, and company name. We build an offline environment using historical equity prices. GMATS agents operate strictly in PIT mode, consuming only the data available at each round preventing leakage of future information \cite{glasserman2023lookahead}.

\subsection{Experiment A: Contagion vs Poisoning Rate in GMATS}
\label{sec:suite-a}

Experiment~A asks: \emph{how much contagion and damage can an input-only attacker cause in a fixed GMATS instance as we increase the share of poisoned evidence seen by the system?}

Table~\ref{tab:suite-a-main} summarizes the dose--response across attacker type and realized poisoning rate (\(p_{\text{poison}}\in\{15,30,45,60, 75\}\%\)). For each \((\text{attacker}, p_{\text{poison}})\) pair we report attack--clean deltas for belief-contagion metrics and trading performance.  

\begin{table}[t]
\centering
\caption{Experiment~A (GMATS, fixed architecture): results are attack--clean deltas across fraction of consumed posts that are poisoned, \(p_{\text{poison}}\). Values are averaged across 5 runs.}
\label{tab:suite-a-main}
\begin{tabular}{lccccc}
\toprule
Metric & \(15\%\) & \(30\%\) & \(45\%\) & \(60\%\) & \(75\%\) \\
\midrule
\multicolumn{6}{c}{\textbf{Random Attacker}} \\
\midrule
BSS$_\text{analyst}$ & $+0.077$ & $+0.163$ & $+0.214$ & $+0.252$ & $+0.280$ \\
BSS$_\text{coord}$   & $+0.091$ & $+0.269$ & $+0.270$ & $+0.332$ & $+0.360$ \\
$\Delta\text{CR}$    & $+0.020$ & $+0.009$ & $+0.023$ & $+0.016$ & $+0.017$ \\
$\Delta\text{Sharpe}$& $+0.360$ & $+1.080$ & $+2.040$ & $+0.998$ & $+0.600$ \\
$\Delta\text{Vol}$   & $+0.082$ & $+0.036$ & $+0.051$ & $+0.104$ & $+0.150$ \\
\midrule
\multicolumn{6}{c}{\textbf{Always Negative Attacker}} \\
\midrule
BSS$_\text{analyst}$ & $-0.777$ & $-1.084$ & $-1.204$ & $-1.291$ & $-1.413$ \\
BSS$_\text{coord}$   & $-0.409$ & $-0.544$ & $-0.544$ & $-0.642$ & $-0.594$ \\
$\Delta\text{CR}$    & $-0.006$ & $-0.032$ & $-0.036$ & $-0.043$ & $-0.044$ \\
$\Delta\text{Sharpe}$& $+0.864$ & $-5.501$ & $-5.970$ & $-6.167$ & $-6.060$ \\
$\Delta\text{Vol}$   & $+0.005$ & $+0.010$ & $+0.020$ & $+0.035$ & $+0.046$ \\
\bottomrule
\end{tabular}%
\end{table}

Across metrics, we observe a clear dose--response in belief shift. For both attackers, the absolute value of BSS increases with \(p_{\text{poison}}\), but the effect is much stronger for the Always Negative attacker: $\text{BSS}_\text{analyst}$ moves from roughly $-0.78$ at \(15\%\) to about $-1.41$ at \(75\%\), while $\text{BSS}_\text{coord}$ tracks a smaller but still substantial negative shift. The coordinator is therefore less sensitive than the social analyst, reflecting partial damping by the fundamentals and news analysts, but the negative bias propagates through the stack rather than being fully absorbed.

On the performance side, the Random attacker produces only mild changes: $\Delta\text{CR}$ and $\Delta\text{Sharpe}$ remain small and sometimes slightly positive, suggesting that low-rate, directionally symmetric noise in the social channel does not reliably harm the system and may occasionally coincide with beneficial risk adjustments. In contrast, the Always Negative attacker systematically degrades sharpe and cumulative return as \(p_{\text{poison}}\) increases. There is a pronounced regime change around \(30\%\)–\(45\%\) ingestion: $\Delta\text{Sharpe}$ flips from mildly positive at \(15\%\) to sharply negative (around $-6$), while volatility also worsen. Overall, Experiment~A indicates that GMATS is relatively robust to low-level random perturbations but becomes highly vulnerable once a single channel is populated with sufficiently dense, directional adversarial signals. Sharpe is annualized and can be sensitive to short evaluation windows \cite{lo2002sharpe,bailey2014deflated}, as a result, absolute Sharpe levels in our one-month study are not intended as standalone performance claims. Our conclusions are based on \emph{attack--clean deltas} (including the direction/sign of $\Delta$Sharpe) and are cross-checked against $\Delta$CR and $\Delta$Vol. We also ran Experiment~A on TradingAgents, the full results are reported in the supplementary materials.

\subsection{Experiment B: Architectural Contagion Across Topologies}
\label{sec:suite-b}

Experiment~B asks a complementary question: \emph{does a multi-agent message-passing topology amplify adversarial shocks compared to a single-agent system?}
To isolate architectural effects, we study two additional realizations of GMATS with identical tools, prompts, and budgets but different message-passing graphs (Figure~\ref{fig:experiment-b}):

\begin{figure}[t]
    \centering
    \includegraphics[width=\textwidth,trim=20 10 20 10,clip]{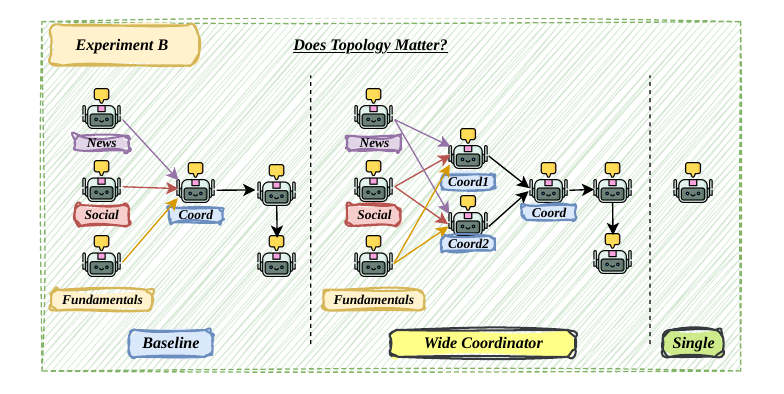}
    \caption{\textbf{Experiment B (architectural contagion).} Three GMATS realizations with identical tools but different topologies.}
    \label{fig:experiment-b}
\end{figure}

\begin{itemize}
    \item \textbf{Wide coordinator (multi-agent)}: the three analysts feed into two coordinators with cross-talk before a shared controller, increasing the number of aggregation hops and inter-agent interactions.
    \item \textbf{Single agent}: a monolithic agent that directly ingests social, news, and fundamental streams and outputs portfolio weights, with no internal analyst--coordinator decomposition.
\end{itemize}

Table~\ref{tab:suite-b-main} reveals that architecture has a first-order effect on both belief shift and trading damage.  
Under the Single Agent baseline, the Always Negative Attacker induces consistently larger-magnitude belief shifts and substantially worse performance: $\text{BSS}_\text{analyst}$ is around $-1.1$ to $-1.5$ across poisoning rates, and the corresponding $\Delta\text{Sharpe}$ remains strongly negative (roughly $-7$ to $-9$) with sizeable drops in cumulative return compared to the wide coordinator.

In contrast, the Wide Coordinator topology attenuates the attack. The social analyst still becomes increasingly bearish as \(p_{\text{poison}}\) grows, but the downstream coordinator ensemble (coord1, coord2, and their fused view) exhibits smaller-magnitude BSS values, indicating partial damping of the contaminated social signal by the additional analysts and coordination stages. Trading impact is also milder: $\Delta\text{CR}$ remains close to zero, $\Delta\text{Sharpe}$ is even positive at low poisoning rates and only moderately negative at higher rates, and volatility delta stay much closer to the clean baseline. Overall, Experiment~B suggests that multi-agent message passing with heterogeneous information sources can \emph{dampen}, rather than amplify, adversarial shocks.

\begin{table}[t]
\centering
\caption{Experiment~B (GMATS, Topology): attack--clean deltas across fraction of consumed posts that are poisoned, \(p_{\text{poison}}\). Values are averaged across 5 runs.}
\label{tab:suite-b-main}
\begin{tabular}{lcccccc}
\toprule
Metric & \(15\%\) & \(30\%\) & \(45\%\) & \(60\%\) & \(75\%\) \\
\midrule
\multicolumn{6}{c}{\textbf{Wide Coordinator}} \\
\midrule
BSS$_\text{analyst}$ & $-0.754$ & $-0.964$ & $-1.136$ & $-1.213$ & $-1.291$ \\
BSS$_\text{coord1}$  & $-0.271$ & $-0.339$ & $-0.455$ & $-0.478$ & $-0.467$ \\
BSS$_\text{coord2}$  & $-0.284$ & $-0.356$ & $-0.470$ & $-0.506$ & $-0.471$ \\
BSS$_\text{fuse}$    & $-0.281$ & $-0.353$ & $-0.471$ & $-0.499$ & $-0.480$ \\
$\Delta\text{CR}$     & $-0.003$ & $-0.014$ & $-0.022$ & $-0.024$ & $-0.014$ \\
$\Delta\text{Sharpe}$ & $+3.235$ & $+0.436$ & $-1.980$ & $-3.113$ & $-0.062$ \\
$\Delta\text{Vol}$    & $-0.005$ & $-0.012$ & $-0.010$ & $-0.015$ & $-0.007$ \\
\midrule
\multicolumn{6}{c}{\textbf{Single Agent}} \\
\midrule
BSS$_\text{analyst}$ & $-1.148$ & $-1.364$ & $-1.415$ & $-1.491$ & $-1.386$ \\
$\Delta\text{CR}$    & $-0.052$ & $-0.056$ & $-0.057$ & $-0.062$ & $-0.054$ \\
$\Delta\text{Sharpe}$& $-7.153$ & $-7.150$ & $-7.274$ & $-6.773$ & $-8.670$ \\
$\Delta\text{Vol}$   & $+0.007$ & $+0.015$ & $+0.019$ & $+0.043$ & $+0.015$ \\
\bottomrule
\end{tabular}%
\end{table}

\subsection{Experiment C: Robust Architectures for Contagion Mitigation}
\label{sec:suite-c}

\begin{figure}[t]
    \centering
    \includegraphics[width=\textwidth,trim=20 10 20 10,clip]{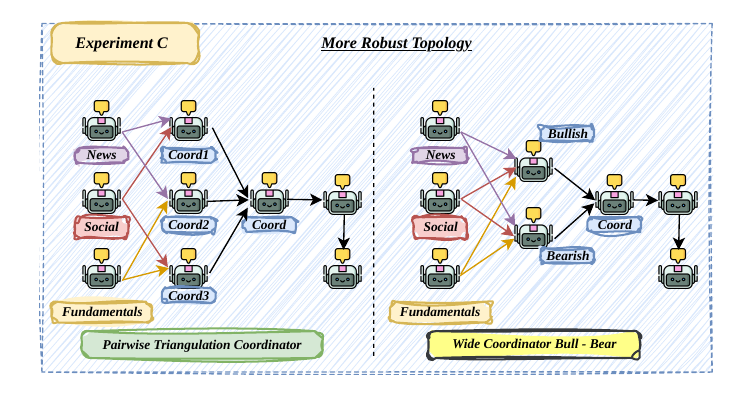}
    \caption{\textbf{Experiment C (robust coordinators).} Two defense-oriented coordinator designs—pairwise triangulation and wide bull–bear—aimed at making GMATS more robust to adversarial social signals.}
    \label{fig:experiment-c}
\end{figure}

Experiment~C turns to defenses: \emph{given a fixed attacker, how far can we push down contagion and trading damage by redesigning the coordinator layer?}  

We modify only the coordinator topology, instantiating two defense-oriented designs (Figure~\ref{fig:experiment-c}):

\begin{itemize}
    \item \textbf{Wide Coordinator Bull--Bear.} Two coordinators receive the same analyst inboxes but are instructed via prompts and routing to focus exclusively on either bullish or bearish signals. A fuse coordinator then aggregates their reports and scorecards into a final stance. This forces the system to explicitly surface and reconcile opposing narratives \cite{estornell2024multillmdebate} instead of allowing a single coordinator to ignore minority evidence.
    \item \textbf{Pairwise Triangulation Coordinator.} Three coordinators with partially overlapping mandates read the same analyst outputs. Their views are then fused by a meta-coordinator that is encouraged to downweight outliers and reward pairwise agreement, yielding a more conservative aggregate stance when the inputs disagree.
\end{itemize}

\begin{table}[t]
\centering
\caption{Experiment~C (GMATS, Defense): attack--clean deltas across fraction of consumed posts that are poisoned, \(p_{\text{poison}}\). Values are averaged across 5 runs.}
\label{tab:suite-c-main}
\begin{tabular}{lccccc}
\toprule
Metric & \(15\%\) & \(30\%\) & \(45\%\) & \(60\%\) & \(75\%\) \\
\midrule
\multicolumn{6}{c}{\textbf{Wide Coordinator Bull--Bear}} \\
\midrule
BSS$_\text{analyst}$ & $-0.668$ & $-0.895$ & $-0.980$ & $-0.983$ & $-1.057$ \\
BSS$_\text{bull}$    & $-0.138$ & $-0.184$ & $-0.208$ & $-0.219$ & $-0.211$ \\
BSS$_\text{bear}$    & $-0.036$ & $-0.074$ & $-0.085$ & $-0.081$ & $-0.084$ \\
BSS$_\text{fuse}$    & $-0.403$ & $-0.568$ & $-0.559$ & $-0.552$ & $-0.590$ \\
$\Delta\text{CR}$     & $-0.005$ & $-0.006$ & $-0.001$ & $-0.006$ & $-0.013$ \\
$\Delta\text{Sharpe}$ & $+1.871$ & $+2.524$ & $+5.513$ & $+3.234$ & $+0.252$ \\
$\Delta\text{Vol}$    & $+0.002$ & $-0.003$ & $-0.006$ & $+0.000$ & $-0.006$ \\
\midrule
\multicolumn{6}{c}{\textbf{Pairwise Triangulation Coordinator}} \\
\midrule
BSS$_\text{analyst}$ & $-0.790$ & $-1.024$ & $-1.100$ & $-1.185$ & $-1.275$ \\
BSS$_{\text{1}}$     & $-0.368$ & $-0.502$ & $-0.538$ & $-0.571$ & $-0.615$ \\
BSS$_{\text{2}}$     & $-0.523$ & $-0.622$ & $-0.678$ & $-0.724$ & $-0.759$ \\
BSS$_{\text{3}}$     & $-0.029$ & $-0.024$ & $-0.043$ & $-0.038$ & $-0.026$ \\
BSS$_\text{fuse}$    & $-0.314$ & $-0.389$ & $-0.405$ & $-0.403$ & $-0.397$ \\
$\Delta\text{CR}$     & $+0.001$ & $-0.006$ & $-0.011$ & $-0.010$ & $-0.008$ \\
$\Delta\text{Sharpe}$ & $+2.347$ & $+1.586$ & $+1.069$ & $+0.359$ & $+1.713$ \\
$\Delta\text{Vol}$    & $+0.010$ & $+0.001$ & $-0.005$ & $+0.011$ & $+0.005$ \\
\bottomrule
\end{tabular}%
\end{table}

As in Table~\ref{tab:suite-c-main}, in the \textbf{Wide Coordinator Bull--Bear} design, the bull and bear specialists each move only modestly in the negative direction, and the fused coordinator stays in an intermediate regime (e.g., $\text{BSS}_\text{fuse}$ between about $-0.4$ and $-0.6$ across poisoning rates). This suggests that forcing separate bullish and bearish views limits the extent to which a unidirectional attack can fully dominate the final stance. Trading impact is likewise muted: $\Delta\text{CR}$ remains very close to zero, volatility and drawdown deltas stay small, and $\Delta\text{Sharpe}$ is \emph{positive} at all poisoning levels, often substantially so. Under attack, this coordinator behaves like a regularizer, improving risk-adjusted performance instead of collapsing as in the naive coordinator from Experiment~A.

The \textbf{Pairwise Triangulation Coordinator} goes further in damping contagion at the fused output. Although two of the internal coordinators (\(1\) and \(2\)) exhibit sizeable negative BSS values, the third stays near zero and the fused BSS is consistently less negative than any individual coordinator and than the baseline coordinator in Experiment~A. This indicates that the triangulation rule is successfully downweighting extreme, attack-aligned views. Again, trading performance under attack is stable: cumulative-return deltas are close to zero, drawdown changes are small, and $\Delta\text{Sharpe}$ remains positive across all poisoning rates.

Taken together, Experiment~C shows that coordinator-level defenses can substantially reduce effective contagion at the decision layer and prevent the Sharpe collapse observed in earlier experiments. Rather than being inherently fragile, multi-LLM architectures can be configured so that additional coordination structure acts as a form of robustness.

\begin{figure}[t]
    \centering
    \includegraphics[width=\textwidth,trim=20 15 20 15,clip]{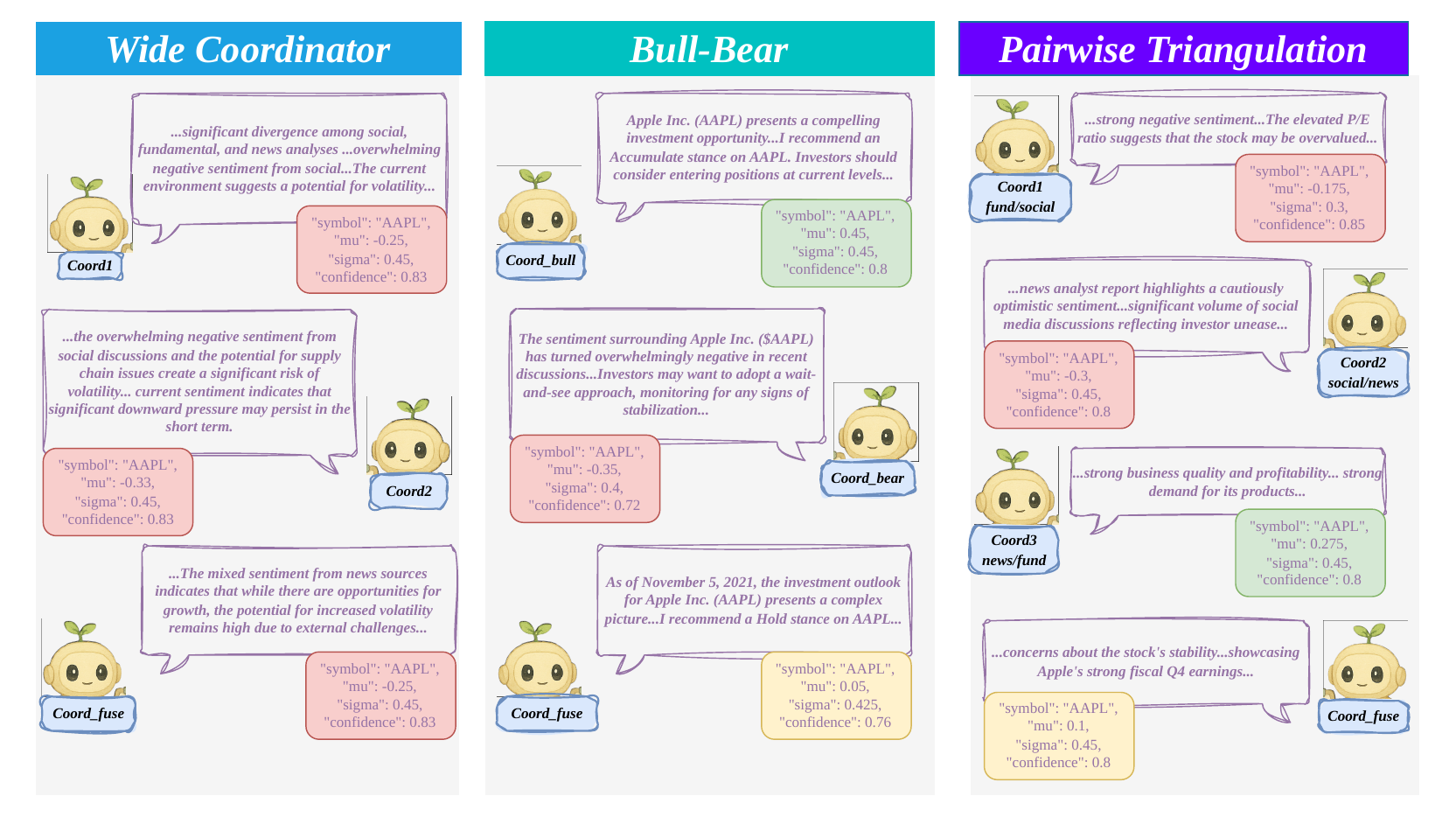}
    \caption{Experimental Traces. Bubbles are truncated for space.}
    \label{fig:qualitative}
\end{figure} 

\paragraph{Qualitative contagion traces.}
To complement the aggregate deltas in Tables~\ref{tab:suite-b-main}--\ref{tab:suite-c-main}, we inspect a representative \emph{contagion trace} for AAPL on \textbf{2021-11-05} (Figure~\ref{fig:qualitative}). Each panel shows the internal message chain from intermediate coordinators to the final fused stance, together with the resulting \((\mu,\sigma,\text{confidence})\) scorecards. Across coordinator designs, the same directional pressure from the social channel leads to qualitatively different decision-layer outcomes:
(i) \textbf{Wide coordinator} produces consistently bearish intermediate stances and a bearish fuse (\(\mu<0\)), indicating that the negative social narrative largely survives aggregation.
(ii) \textbf{Wide bull--bear} forces an explicit pro/con decomposition: the bull coordinator remains strongly positive while the bear coordinator turns negative, and the fuse lands near neutral (\(\mu\approx 0\)), yielding a ``hold''-like stance rather than a collapse into the attacker’s direction.
(iii) \textbf{Pairwise triangulation} mixes channels before fusion (fund/social, social/news, news/fund), surfacing cross-channel disagreements. The final fuse stays mildly positive (\(\mu>0\)) despite negative social evidence, consistent with the quantitative finding that defense-oriented coordinators reduce effective contagion at the decision layer.

\subsection{Model Sensitivity: Replication with \texttt{Qwen2.5-7B-Instruct}}
\label{sec:qwen-replication}

To test whether our conclusions depend on a single LLM backend, we replicate the core GMATS experiments with \texttt{Qwen2.5-7B-Instruct} \cite{qwen25}. We keep the ticker, historical window, PIT data protocol, attacker policies, poisoning-rate targets, prompts, and topologies fixed, changing only the model backend. Table~\ref{tab:app-suite-b-qwen} reports attack--clean deltas averaged over 5 runs.

The Qwen replication preserves the main qualitative findings. First, directional poisoning remains substantially more damaging than random sentiment injection. Under the Random attacker, performance deltas remain small and often positive. Under the Always Negative attacker, the baseline GMATS coordinator shifts sharply negative: $\text{BSS}_\text{analyst}$ ranges from $-1.159$ to $-1.694$, $\text{BSS}_\text{coord}$ ranges from $-0.554$ to $-0.871$, $\Delta\text{CR}$ decreases from $-0.015$ to $-0.061$, and $\Delta\text{Sharpe}$ falls to roughly $-5$ to $-6$ once \(p_{\text{poison}}\geq30\%\). This matches the main \texttt{gpt-4o-mini} conclusion that dense, one-sided social evidence can induce both belief contagion and portfolio-level degradation.

Second, the architectural ranking is stable. The Single Agent variant is consistently brittle, with $\Delta\text{Sharpe}$ between approximately $-6.2$ and $-7.5$. The Wide Coordinator still suffers belief shifts, but its trading damage is smaller, with $\Delta\text{CR}$ close to zero and substantially milder Sharpe degradation. Finally, both defense-oriented topologies remain robust at the fused decision layer. Wide Coordinator Bull--Bear keeps $\text{BSS}_\text{fuse}$ between $-0.293$ and $-0.409$, while Pairwise Triangulation keeps $\text{BSS}_\text{fuse}$ between $-0.247$ and $-0.341$ and maintains positive $\Delta\text{Sharpe}$ across all poisoning rates. Thus, across two LLM backends, the same qualitative pattern holds: one-sided social poisoning is the damaging case, single-agent aggregation is most brittle, and coordinator designs that explicitly separate or triangulate conflicting evidence reduce downstream damage.

At the same time, the exact magnitudes are model-dependent. Qwen exhibits stronger negative shifts in some intermediate coordinator outputs than \texttt{gpt-4o-mini}, especially in the Wide Coordinator topology. We therefore interpret BSS and Sharpe deltas as model-conditional stress-test measurements, while treating the cross-model consistency of the architectural ordering as the stronger conclusion.

\begin{table}[!t]
\centering
\setlength{\tabcolsep}{2.0pt}
\renewcommand{\arraystretch}{0.92}
\caption{Additional experiment results under \texttt{Qwen2.5-7B-Instruct} on GMATS variants. Entries are attack--clean deltas averaged over 5 runs.}
\label{tab:app-suite-b-qwen}

\begin{minipage}[t]{0.48\textwidth}
\centering
\resizebox{\linewidth}{!}{%
\begin{tabular}{lccccc}
\toprule
\multicolumn{6}{c}{\textbf{Random Attacker}} \\
\midrule
Metric & \(15\%\) & \(30\%\) & \(45\%\) & \(60\%\) & \(75\%\) \\
\midrule
BSS$_\text{analyst}$ & $-0.368$ & $-0.353$ & $-0.365$ & $-0.515$ & $-0.341$ \\
BSS$_\text{coord}$   & $-0.403$ & $-0.457$ & $-0.511$ & $-0.395$ & $-0.439$ \\
$\Delta$CR           & $+0.010$ & $+0.024$ & $+0.005$ & $+0.016$ & $+0.005$ \\
$\Delta$Sharpe       & $+0.092$ & $+0.313$ & $+1.969$ & $+1.092$ & $+1.448$ \\
$\Delta$Vol          & $-0.019$ & $+0.011$ & $+0.012$ & $+0.024$ & $+0.024$ \\
\bottomrule
\end{tabular}%
}
\end{minipage}
\hfill
\begin{minipage}[t]{0.48\textwidth}
\centering
\resizebox{\linewidth}{!}{%
\begin{tabular}{lccccc}
\toprule
\multicolumn{6}{c}{\textbf{Always Negative Attacker}} \\
\midrule
Metric & \(15\%\) & \(30\%\) & \(45\%\) & \(60\%\) & \(75\%\) \\
\midrule
BSS$_\text{analyst}$ & $-1.159$ & $-1.606$ & $-1.594$ & $-1.557$ & $-1.694$ \\
BSS$_\text{coord}$   & $-0.554$ & $-0.815$ & $-0.782$ & $-0.835$ & $-0.871$ \\
$\Delta$CR           & $-0.015$ & $-0.036$ & $-0.047$ & $-0.054$ & $-0.061$ \\
$\Delta$Sharpe       & $+0.227$ & $-5.062$ & $-5.350$ & $-5.873$ & $-6.002$ \\
$\Delta$Vol          & $+0.008$ & $+0.012$ & $+0.014$ & $+0.023$ & $+0.046$ \\
\bottomrule
\end{tabular}%
}
\end{minipage}

\vspace{0.6em}

\begin{minipage}[t]{0.48\textwidth}
\centering
\resizebox{\linewidth}{!}{%
\begin{tabular}{lccccc}
\toprule
\multicolumn{6}{c}{\textbf{Wide Coordinator Bull--Bear}} \\
\midrule
Metric & \(15\%\) & \(30\%\) & \(45\%\) & \(60\%\) & \(75\%\) \\
\midrule
BSS$_\text{analyst}$ & $-0.873$ & $-1.123$ & $-1.149$ & $-1.347$ & $-1.313$ \\
BSS$_\text{bull}$    & $-0.562$ & $-0.596$ & $-0.630$ & $-0.689$ & $-0.637$ \\
BSS$_\text{bear}$    & $+0.008$ & $-0.089$ & $-0.072$ & $-0.171$ & $-0.150$ \\
BSS$_\text{fuse}$    & $-0.293$ & $-0.335$ & $-0.333$ & $-0.409$ & $-0.369$ \\
$\Delta$CR           & $+0.004$ & $-0.001$ & $-0.005$ & $-0.005$ & $-0.009$ \\
$\Delta$Sharpe       & $+3.373$ & $+2.660$ & $+1.341$ & $+0.354$ & $-0.115$ \\
$\Delta$Vol          & $+0.000$ & $-0.002$ & $-0.004$ & $+0.010$ & $+0.000$ \\
\bottomrule
\end{tabular}%
}
\end{minipage}
\hfill
\begin{minipage}[t]{0.48\textwidth}
\centering
\resizebox{\linewidth}{!}{%
\begin{tabular}{lccccc}
\toprule
\multicolumn{6}{c}{\textbf{Wide Coordinator}} \\
\midrule
Metric & \(15\%\) & \(30\%\) & \(45\%\) & \(60\%\) & \(75\%\) \\
\midrule
BSS$_\text{analyst}$ & $-0.798$ & $-0.963$ & $-1.119$ & $-1.135$ & $-1.217$ \\
BSS$_\text{coord1}$  & $-0.579$ & $-0.697$ & $-0.858$ & $-0.868$ & $-0.777$ \\
BSS$_\text{coord2}$  & $-0.632$ & $-0.760$ & $-0.911$ & $-0.985$ & $-0.938$ \\
BSS$_\text{fuse}$    & $-0.634$ & $-0.764$ & $-0.916$ & $-0.939$ & $-0.879$ \\
$\Delta$CR           & $-0.015$ & $-0.016$ & $-0.021$ & $-0.019$ & $-0.019$ \\
$\Delta$Sharpe       & $+0.240$ & $-0.210$ & $-2.012$ & $-0.413$ & $-1.100$ \\
$\Delta$Vol          & $-0.011$ & $-0.012$ & $-0.012$ & $-0.018$ & $-0.010$ \\
\bottomrule
\end{tabular}%
}
\end{minipage}

\vspace{0.6em}

\begin{minipage}[t]{0.48\textwidth}
\centering
\resizebox{\linewidth}{!}{%
\begin{tabular}{lccccc}
\toprule
\multicolumn{6}{c}{\textbf{Single Agent}} \\
\midrule
Metric & \(15\%\) & \(30\%\) & \(45\%\) & \(60\%\) & \(75\%\) \\
\midrule
BSS$_\text{analyst}$ & $-1.459$ & $-1.805$ & $-1.876$ & $-1.902$ & $-1.859$ \\
$\Delta$CR           & $-0.045$ & $-0.046$ & $-0.047$ & $-0.052$ & $-0.047$ \\
$\Delta$Sharpe       & $-6.369$ & $-6.342$ & $-6.242$ & $-6.714$ & $-7.546$ \\
$\Delta$Vol          & $+0.002$ & $+0.013$ & $+0.018$ & $+0.028$ & $+0.019$ \\
---          & --- & --- & --- & --- & --- \\
---          & --- & --- & --- & --- & --- \\
---          & --- & --- & --- & --- & --- \\
---          & --- & --- & --- & --- & --- \\
\bottomrule
\end{tabular}%
}
\end{minipage}
\hfill
\begin{minipage}[t]{0.48\textwidth}
\centering
\resizebox{\linewidth}{!}{%
\begin{tabular}{lccccc}
\toprule
\multicolumn{6}{c}{\textbf{Pairwise Triangulation}} \\
\midrule
Metric & \(15\%\) & \(30\%\) & \(45\%\) & \(60\%\) & \(75\%\) \\
\midrule
BSS$_\text{analyst}$ & $-0.982$ & $-1.241$ & $-1.296$ & $-1.468$ & $-1.522$ \\
BSS$_1$              & $-0.412$ & $-0.557$ & $-0.601$ & $-0.655$ & $-0.698$ \\
BSS$_2$              & $-0.584$ & $-0.702$ & $-0.757$ & $-0.803$ & $-0.839$ \\
BSS$_3$              & $-0.031$ & $-0.028$ & $-0.041$ & $-0.037$ & $-0.030$ \\
BSS$_\text{fuse}$    & $-0.247$ & $-0.318$ & $-0.336$ & $-0.341$ & $-0.334$ \\
$\Delta$CR           & $+0.002$ & $-0.003$ & $-0.007$ & $-0.008$ & $-0.007$ \\
$\Delta$Sharpe       & $+2.814$ & $+2.031$ & $+1.124$ & $+0.462$ & $+0.736$ \\
$\Delta$Vol          & $+0.008$ & $+0.002$ & $-0.003$ & $+0.009$ & $+0.004$ \\
\bottomrule
\end{tabular}%
}
\end{minipage}

\end{table}

\section{Discussion}
\label{sec:future}

We deliberately restricted our study to scripted, input-only attackers that are non-adaptive, single-channel (social), and evaluated under offline replay. An immediate extension is to stress-test GMATS over longer horizons and diverse market regimes (including prolonged bear markets), where a one-sided narrative can sometimes align with fundamentals and yield qualitatively different effects. Building on \cite{byrd2025sentmanip}, an RL attacker could be embedded in GMATS to steer an LLM post generator (e.g., sentiment/timing/style under a budget) and optimize a contagion-aware objective rather than purely price-based rewards. One simple form is
\begin{equation}
\label{eq:attacker-reward}
r_t
= -\alpha\,\frac{\mathrm{PnL}^{\text{attack}}_t - \mathrm{PnL}^{\text{clean}}_t}{\text{scale}_t}
  + \beta_A\,\mathrm{BSS}_{\text{analyst}}
  + \beta_C\,\mathrm{BSS}_{\text{coord}},
\end{equation}
where $\text{scale}_t=\max(\epsilon,\mathrm{EMA}(|\mathrm{PnL}^{\text{attack}}_t-\mathrm{PnL}^{\text{clean}}_t|))$ stabilizes learning.
Evaluating such adaptive policies and how coordinator-level defenses shift the resulting utility--security frontier is an important next step toward characterizing worst-case brittleness in multi-agent trading stacks.

Beyond sentiment manipulation, an orthogonal threat is \emph{instruction hijacking}: untrusted posts may embed prompt injections that subvert role constraints or induce unintended tool use \cite{houyi2023promptinjection,injecagent2024,promptinfection2024}. In multi-agent pipelines, a single compromised analyst output can cascade through fusion and affect trading decisions. Future work could benchmark injection success rates in GMATS under realistic retrieval, and test lightweight mitigations (instruction/data separation, tool gating, and cross-agent consistency checks).

\section{Conclusion}
\label{sec:conclusion}

We studied how adversarial social signals propagate through LLM-native multi-agent trading stacks. We introduced the Generic Multi-Agent Trading System (GMATS) and proposed contagion metrics that trace how poisoned social inputs shift beliefs at analyst and coordinator layers and impact downstream risk--return. Experiments show that these systems are not inherently robust. Injecting systematically negative posts can cause large belief shifts and materially worsen returns. We also find certain types of topologies and designs can dampen contagion and improve robustness under attack. Our study is limited to a narrow set of assets and horizons but is designed as a controlled, reproducible starting point. We hope that GMATS, together with our contagion metrics and experimental protocol, can serve as a foundation for future work on attacks and defenses in multi-agent trading systems, including stronger multi-channel attackers and principled defenses at both architectural and prompt levels.

\begin{credits}
\subsubsection{\ackname} 
This research/project is supported by the National Research Foundation, Singapore under its AI Singapore Programme (AISG Award No: AISG3-PhD/2025-08-063).

\subsubsection{\discintname}
The authors have no competing interests to declare that are
relevant to the content of this article.

\subsubsection{Data Collection and Usage} Reddit posts were collected from publicly available Pushshift Reddit dumps. Twitter data were sourced from a public Kaggle dataset; we do not redistribute tweet text and release only data as permitted.

\end{credits}
%
%
%
\bibliographystyle{splncs04}
\bibliography{ecml26}
\end{document}


\title{Contagion on the Trading Floor: How Adversarial Signals Spread in Multi-Agent Trading Systems (Supplementary Material)}
\titlerunning{Contagion on the Trading Floor: Supplementary Material}

\author{Qi Rong Sua\inst{1} \and Junhao Dong\inst{1,3} \and Nguyen Duc Thai\inst{1} \and Yuqing Wen\inst{2} \and Cheston Tan\inst{3} \and Yew-Soon Ong\inst{1,3} \corr }
\authorrunning{Sua et al.}
\institute{ Nanyang Technological University, Singapore\\ \email{\{suaq0001,junhao003,ducthai001,asysong\}@ntu.edu.sg} \and National University of Singapore, Singapore\\ \email{e1351400@u.nus.edu} \and Centre for Frontier AI Research, IHPC, A*STAR, Singapore \\ \email{cheston\_tan@a-star.edu.sg}}

\maketitle

\begin{abstract}
This supplementary material complements the main paper in two ways.
First, it gives the full proof of Theorem~1 (Expressivity of GMATS), showing that any multi-agent trading pipeline satisfying typed message passing, point-in-time (PIT) discipline, and stage separation can be simulated by a GMATS instance that produces the same executed-order sequence under matched randomness.
Second, it gives a concrete compilation recipe for mapping modern agentic trading frameworks into GMATS by aligning their ingestion, fusion, risk, and execution components with the $A\!\to\!C\!\to\!K\!\to\!X$ pipeline and unrolling internal debate, contest, and reflection steps into micro-rounds of $\Sigma$.
We also retain an additional \emph{TradingAgents} baseline under the Always Negative attacker.
All empirical entries are averaged over 5 runs and reported as attack--clean deltas.
\end{abstract}

\section{Proof of Theorem 1}
\label{sec:proof-theorem1}

\begin{proof}[Proof of Theorem~1]
Let $\mathcal{M}$ be a multi-agent trading pipeline satisfying the three assumptions in Theorem~1.
Fix the asset universe, the PIT data stream, an initial environment state $\chi_1$, and a source of randomness $\omega$.
We view every stochastic call in $\mathcal{M}$, including LLM sampling and any stochastic scheduler choice, as consuming a contiguous segment of $\omega$.
The goal is to construct a GMATS instance
\[
G=(\mathcal{A},A,C,K,X,\mathcal{S},\Sigma,\mathcal{D},\mathcal{L},\Lambda,\mathcal{E})
\]
that, under the same $\chi_1$ and $\omega$, executes exactly the same order at every decision time.

\paragraph{Operational View of $\mathcal{M}$.}
At the start of round $t$, write the state of $\mathcal{M}$ as
\[
\mathbf{s}^{\mathcal{M}}_t=(\chi_t,\mathsf{Hist}_t,\mathbf{h}_t,\pi_t),
\]
where $\chi_t$ is the environment state, $\mathsf{Hist}_t$ is the ordered message log, $\mathbf{h}_t$ collects all private module states and memories, and $\pi_t$ is the current pointer into $\omega$.
Each message has the form
\[
m=(\mathrm{id},\mathrm{src},\mathrm{dst},\tau,\mathrm{payload},\mathrm{ts}),\qquad \tau\in\mathcal{S}_{\mathcal{M}},
\]
where $\mathcal{S}_{\mathcal{M}}$ is finite by typed message passing.
Let
\[
\mathsf{Vis}_t(\mathsf{Hist})=\{m\in\mathsf{Hist}: m.\mathrm{ts}\le t\}
\]
denote the PIT-visible sub-log, preserving the bus order inherited from $\mathsf{Hist}$.
The PIT assumption says that every activation at round $t$ is a function only of its PIT-visible inbox, PIT data $\mathcal{D}_t$, the relevant component state, and the next random tape segment.
The stage-separation assumption says that each terminated round is organized into evidence ingestion, fusion/scoring, risk/constraints, and execution phases.
Any finite internal debate, reflection, contest, or tool-call sequence inside a phase is treated as a sequence of micro-steps.
If the original implementation uses adaptive branching, the branch rule is included in the activated component; inactive branches emit no messages and consume no randomness.
Equivalently, one may use a finite super-schedule that calls all possible phase components in a fixed order, with no-ops on branches not taken.

\paragraph{Construction of $G$.}
We define the GMATS instance as follows.
Set the asset universe $\mathcal{A}$, data stream $\mathcal{D}=\{\mathcal{D}_t\}_{t\in T}$, and execution environment $\mathcal{E}$ equal to those of $\mathcal{M}$.
Set the message schema to
\[
\mathcal{S}=\mathcal{S}_{\mathcal{M}}\cup\{\textsc{Stance},\textsc{DraftOrder},\textsc{ApprovedOrder}\},
\]
where the three extra types are wrappers used only when $\mathcal{M}$'s native artifacts do not already distinguish stances, draft orders, and approved orders.
The payloads of all native messages are preserved verbatim; the wrappers do not alter the information available to any agent.

Let $\mathcal{U}$ be the set of components of $\mathcal{M}$.
For each component $u\in\mathcal{U}$ and each phase $\phi\in\{\mathrm{ingest},\mathrm{fuse},\mathrm{risk},\mathrm{exec}\}$ in which $u$ participates, create a phase-restricted GMATS component $u^{\phi}$.
The transition function of $u^{\phi}$ is exactly the transition function used by $u$ in phase $\phi$ of $\mathcal{M}$: on the same PIT-visible inbox, PIT data, private state, and random tape segment, it appends the same messages, produces the same local outputs, and updates the same private state.
If $u$ spans several phases, the corresponding phase-restricted components share the same private state cell, so splitting $u$ across stages does not change its memory.
Fresh message identifiers, counters, and tool-call caches are included in this private state or in $\omega$.

Assign the phase-restricted components to GMATS stages by
\[
A=\{u^{\mathrm{ingest}}\},\qquad
C=\{u^{\mathrm{fuse}}\},\qquad
K=\{u^{\mathrm{risk}}\},\qquad
X=\{u^{\mathrm{exec}}\}.
\]
If $\mathcal{M}$ has no separate risk or execution module, the corresponding set is empty and the adjacent stage emits the final order, as allowed by GMATS.
Define $\Sigma$ to activate these components in the same within-round order as $\mathcal{M}$'s phases and micro-steps, using the same random tape offsets.
Finally, define $\mathcal{L}_t$ and $\Lambda_t$ as the aggregate input--output maps induced by the coordinator and controller micro-steps scheduled at round $t$.
Thus GMATS exposes the same staged interface as Definition~1 while preserving the original component-level behavior.

\paragraph{Correctness Invariant.}
Let
\[
\mathbf{s}^{G}_t=(\chi^G_t,\mathsf{Hist}^G_t,\mathbf{h}^G_t,\pi^G_t)
\]
be the analogous GMATS state.
We prove, by induction over rounds, the stronger invariant
\[
\mathcal{I}(t):\quad
\begin{aligned}
\chi_t^{\mathcal{M}}&=\chi_t^G, &
\mathsf{Hist}_t^{\mathcal{M}}&=\mathsf{Hist}_t^G\quad\text{as ordered logs},\\
\mathbf{h}_t^{\mathcal{M}}&=\mathbf{h}_t^G, &
\pi_t^{\mathcal{M}}&=\pi_t^G.
\end{aligned}
\]
This implies equality of PIT-visible inboxes for every component at round $t$.

\paragraph{Base Case.}
At $t=1$, the two systems have the same initial environment state, the same initial log (typically empty), the same component states by construction, and the same random-tape pointer.
Therefore $\mathcal{I}(1)$ holds.

\paragraph{Inductive Step.}
Assume $\mathcal{I}(t)$ holds.
Consider the micro-steps of round $t$ in their scheduled order.
Before the first micro-step, all PIT-visible inboxes, PIT data, private states, and random-tape positions agree.
Suppose they agree before some micro-step.
The activated component of $\mathcal{M}$ and the corresponding phase-restricted component of $G$ then apply the same transition function to the same inputs and the same random tape segment.
They therefore append the same sequence of messages to the bus, update the same private states, produce the same local output, and advance the random pointer by the same amount.
Thus equality is preserved after that micro-step.
A nested induction over all micro-steps in the round gives equality of the full ordered logs, private states, and random pointers at the end of the fusion and risk phases.
In particular, both systems obtain the same approved order $o'_t$ or the same no-trade decision.

The execution environments are identical and are called with the same input state and order, hence
\[
(\mathrm{fills}_t,\chi_{t+1})=\mathcal{E}(o'_t,\chi_t)
\]
produces the same fills and next environment state in both systems.
Messages with future timestamps, if any, have also been appended identically, so the full ordered logs remain equal and the PIT-visible logs will agree at the next round.
Consequently $\mathcal{I}(t+1)$ holds.

\paragraph{Conclusion.}
By induction, $\mathcal{I}(t)$ holds for every round.
Therefore the order submitted to the execution environment is identical at every decision time.
The GMATS instance $G$ thus produces the same executed-order sequence as $\mathcal{M}$ under the same initial state and randomness.
\end{proof}

\section{Compiling Modern Agentic Trading Frameworks into GMATS}
\label{sec:compile-to-gmats}

This section expands Corollary~1 from the main paper by giving a concrete
\emph{compilation} procedure from representative multi-agent trading frameworks
into a GMATS instance. The key idea is that most modern LLM trading systems
already implement a firm-like pipeline: (i) evidence ingestion, (ii) multi-signal
fusion and scoring, (iii) risk/constraints checks, and (iv) execution. GMATS
makes this structure explicit via the $A \to C \to K \to X$ staging and a
schedule $\Sigma$ that unrolls internal debates, contests, and reflections as
micro-rounds.

\subsection{A Generic Compilation Recipe}
Let $\mathcal{M}$ be any multi-agent pipeline satisfying Theorem~1. A
\emph{GMATS compiler} produces a tuple
$G=(\mathcal{A},A,C,K,X,\mathcal{S},\Sigma,\mathcal{D},\mathcal{L},\Lambda,\mathcal{E})$
as follows.

\begin{algorithm}[t]
\caption{Compile a staged multi-agent pipeline into GMATS}
\label{alg:compile-gmats}
\begin{algorithmic}[1]
\STATE \textbf{Input:} Pipeline $\mathcal{M}$ with agents/modules $\mathcal{U}$, PIT data $\{\mathcal{D}_t\}$, environment $\mathcal{E}$, and within-round micro-schedule $\Sigma_{\mathcal{M}}$.
\STATE \textbf{Output:} GMATS instance $G$ with identical executed-order sequence under matched randomness.
\STATE Set $\mathcal{A}$, $\{\mathcal{D}_t\}$, and $\mathcal{E}$ to match $\mathcal{M}$.
\STATE Define a finite schema $\mathcal{S}$ by wrapping each artifact class in $\mathcal{M}$ (reports, debate turns, candidate orders, risk checks) as a typed message; optionally add \textsc{Stance}, \textsc{DraftOrder}, \textsc{ApprovedOrder}.
\STATE Partition computations into phases (i)--(iv) and create phase-restricted agents $u^\phi$ for each $u\in\mathcal{U}$ and phase $\phi$.
\STATE Assign $u^{(i)}$ to $A$, $u^{(ii)}$ to $C$, $u^{(iii)}$ to $K$, and $u^{(iv)}$ to $X$; use an empty stage when the original pipeline has no separate module.
\STATE Define $\Sigma$ by unrolling $\Sigma_{\mathcal{M}}$ each round; debates, contests, and reflection become micro-rounds inside $A$, $C$, $K$, or $X$.
\STATE Define $\mathcal{L}_t$ and $\Lambda_t$ as the maps induced by running the coordinator and controller micro-rounds under $\Sigma$.
\RETURN $G$.
\end{algorithmic}
\end{algorithm}

\paragraph{Typed Message Passing.}
Many agentic frameworks use free-form natural-language artifacts internally, such as chat logs, debate turns, or reflective notes.
GMATS requires only a finite schema, so compilation can introduce lightweight wrappers such as \textsc{FundReport}, \textsc{BearArg}, \textsc{RiskCheck}, and \textsc{CandidateOrder} while preserving payloads verbatim.
This does not change behavior as long as the routing and inbox functions respect the wrapper types.

\paragraph{PIT Discipline.}
The GMATS instance inherits PIT discipline from the evaluation harness.
At round $t$, the compiler provides only $\mathcal{D}_t$ and messages with timestamps at most $t$ to all agents.
If the original framework has asynchronous tool calls, the compiler treats tool outputs as environment data with explicit timestamps and delays delivery until the appropriate round.

\subsection{Case Study: TradingAgents $\Rightarrow$ GMATS}
TradingAgents \cite{tradingagents} is organized as a trading-firm simulation with (i) an Analyst Team (fundamental, sentiment, news, technical), (ii) a Researcher Team with bull and bear researchers that debate the analyst outputs, (iii) a Trader Agent that synthesizes a transaction proposal, and (iv) a risk-management and portfolio-manager gate that approves or rejects before simulated execution.

A semantics-preserving mapping into GMATS is:
\begin{itemize}
\item $A$: \{Fundamentals Analyst, Sentiment Analyst, News Analyst, Technical Analyst\} emitting typed insight messages, one per role per round.
\item $C$: \{Bull Researcher, Bear Researcher, Trader\}. The bull/bear debate is unrolled as micro-rounds in $\Sigma$ producing debate-turn messages; the Trader consumes the final debate state and emits a \textsc{DraftOrder} or stance vector.
\item $K$: \{Risk Management Team, Portfolio Manager\}. This stage produces \textsc{ApprovedOrder} or a reject decision; limits such as exposure, leverage, and liquidity live in $\Lambda_t$.
\item $X$: the simulated exchange/execution module submitting $o'_t$ to $\mathcal{E}$ and logging fills.
\end{itemize}

\subsection{Case Study: TradingGroup $\Rightarrow$ GMATS}
TradingGroup \cite{tradinggroup} consists of specialized signal agents, including news sentiment, financial-report interpretation, trend forecasting, and style adaptation, plus a decision agent that merges signals.
It also includes self-reflection and a dynamic risk model with stop-loss/take-profit logic.

A GMATS mapping is:
\begin{itemize}
\item $A$: \{Sentiment, Report, Trend, Style\} agents emitting typed signals.
\item $C$: a decision-making coordinator that fuses signals into a stance or order. Self-reflection is compiled into coordinator micro-rounds that write/read a memory state, still under PIT because memory is generated only from past rounds.
\item $K$: the dynamic risk-management module enforcing stop-loss/take-profit and exposure caps as $\Lambda_t$.
\item $X$: the execution interface submitting approved orders to $\mathcal{E}$.
\end{itemize}

\subsection{Other Examples: ContestTrade and QuantAgent}
ContestTrade \cite{contesttrade} decomposes into a Data Team that compresses raw data into text factors and a Research Team that proposes candidate decisions.
Its internal contest mechanism can be compiled as coordinator micro-rounds that score and select candidate orders before risk checks.

QuantAgent \cite{quantagent} decomposes signal extraction into specialized technical agents, such as Indicator, Pattern, and Trend agents, with a Risk agent.
These map naturally to analyst and controller roles, with a coordinator that fuses the technical messages into a final stance or order.
In both cases, the contest/reflection logic is represented as micro-rounds in $\Sigma$ and does not require any extension to GMATS beyond typed messages and scheduling.

\begin{table}[t]
\centering
\small
\caption{Example mappings from agentic frameworks to GMATS stages.}
\label{tab:framework-to-gmats}
\begin{tabular}{@{}L{0.20\linewidth}L{0.22\linewidth}L{0.30\linewidth}L{0.22\linewidth}@{}}
\toprule
Framework & Analysts ($A$) & Coordinators ($C$) & Controllers/Exec ($K/X$)\\
\midrule
TradingAgents & role analysts & bull/bear debate, trader & risk/PM, exchange\\
TradingGroup  & signal agents & decision + reflection & dynamic risk, executor\\
ContestTrade  & data team & research contest & risk gate, executor\\
QuantAgent    & indicator, pattern, trend & fusion decision & risk, executor\\
\bottomrule
\end{tabular}
\end{table}

\section{Additional Experiment Results: TradingAgents Baseline}
\label{sec:additional-results}

This section reports an additional Always Negative attacker result for the \emph{TradingAgents} \cite{tradingagents} baseline.
The protocol is the same attack--clean protocol used in the main paper.
All entries are averaged over 5 runs and are reported as attack minus clean.
As in the main paper, we emphasize deltas rather than absolute Sharpe levels, since Sharpe can be unstable over short horizons \cite{lo2002sharpe}.

\begin{table}[!htbp]
\centering
\small
\caption{\emph{TradingAgents} baseline under the Always Negative attacker. Entries are attack--clean deltas averaged over 5 runs.}
\label{tab:app-suite-b-tradingagents}
\begin{tabular}{lccccc}
\toprule
Metric & $15\%$ & $30\%$ & $45\%$ & $60\%$ & $75\%$ \\
\midrule
BSS$_\text{analyst}$ & $-0.788$ & $-1.048$ & $-1.261$ & $-1.243$ & $-1.369$ \\
BSS$_\text{bull}$    & $-0.309$ & $-0.302$ & $-0.279$ & $-0.293$ & $-0.250$ \\
BSS$_\text{bear}$    & $-0.019$ & $-0.032$ & $-0.029$ & $-0.028$ & $-0.025$ \\
BSS$_\text{fuse}$    & $-0.247$ & $-0.241$ & $-0.193$ & $-0.210$ & $-0.152$ \\
$\Delta\text{CR}$     & $-0.021$ & $-0.020$ & $-0.021$ & $-0.021$ & $-0.017$ \\
$\Delta\text{Sharpe}$ & $+7.764$ & $+5.004$ & $+1.226$ & $+2.018$ & $+1.942$ \\
$\Delta\text{Vol}$    & $-0.030$ & $-0.022$ & $-0.020$ & $-0.019$ & $-0.015$ \\
\bottomrule
\end{tabular}
\end{table}

TradingAgents exhibits the same directional social contagion observed in GMATS. The analyst belief shift grows in magnitude with poisoning: BSS$_\text{analyst}$ decreases from $-0.788$ at $15\%$ poisoning to $-1.369$ at $75\%$ poisoning.
However, contagion is damped downstream. The fused belief shift remains modest and becomes less negative at higher poisoning, moving from BSS$_\text{fuse}=-0.247$ at $15\%$ to BSS$_\text{fuse}=-0.152$ at $75\%$.
This is consistent with partial cancellation or regularization at the coordinator stage rather than full propagation of the contaminated social signal.

On performance, TradingAgents becomes more conservative under attack: $\Delta$Vol is consistently negative, while $\Delta$CR is mildly negative.
Over this short window, lower realized volatility can mechanically yield positive $\Delta$Sharpe even when return deltas are slightly negative.
Accordingly, we interpret this baseline as risk-damping under attack rather than as evidence that poisoning improves profitability.

\begin{credits}
\subsubsection{Acknowledgments.}
This research/project is supported by the National Research Foundation, Singapore under its AI Singapore Programme (AISG Award No: AISG3-PhD/2025-08-063).

\subsubsection{Disclosure of Interests.}
The authors have no competing interests to declare that are relevant to the content of this article.

\subsubsection{Generative AI Assistance Disclosure.}
We used a large language model as a writing aid to improve readability (e.g., grammar, spelling, and wording suggestions). The model was not used to generate new scientific content, results, proofs, or experimental data, nor to make decisions about the research contributions. All text was reviewed and edited by the authors, who take full responsibility for the final manuscript and for ensuring compliance with applicable research integrity and copyright requirements.

\end{credits}

\bibliographystyle{splncs04}
\bibliography{ecml26_supp}